\documentclass[11pt]{article}

\usepackage[preprint]{acl}

\usepackage{times}
\usepackage{latexsym}
\usepackage{amsmath}
\usepackage{amssymb} 
\usepackage{subcaption}
\usepackage[T1]{fontenc}

\usepackage{booktabs}
\usepackage{tabularx}
\usepackage{threeparttable}
\usepackage[utf8]{inputenc}

\usepackage{xurl}

\usepackage{microtype}

\usepackage{inconsolata}

\usepackage{graphicx}

\usepackage{listings}
\usepackage{xcolor}
\usepackage{hyperref}

\definecolor{ghbg}{HTML}{F6F8FA}
\definecolor{ghcomment}{HTML}{6A737D}
\definecolor{ghkeyword}{HTML}{D73A49}
\definecolor{ghstring}{HTML}{032F62}
\definecolor{ghidentifier}{HTML}{24292E}

\lstdefinestyle{github}{
    language=Python,
    backgroundcolor=\color{ghbg},
    basicstyle=\ttfamily\small\color{ghidentifier},
    keywordstyle=\color{ghkeyword}\bfseries,
    stringstyle=\color{ghstring},
    commentstyle=\color{ghcomment}\itshape,
    showstringspaces=false,
    breaklines=true,
    frame=single,
    rulecolor=\color{black!10},
    frameround=tttt,
    xleftmargin=4pt,
    xrightmargin=4pt,
    aboveskip=6pt,
    belowskip=6pt,
}

\title{CLT-Forge: A Scalable Library for
Cross-Layer Transcoders and Attribution Graphs}
\author{
\textbf{
Florent Draye$^{1}$, 
Vedant Palit$^{2}$,
Abir Harrasse$^{2}$,
Tung-Yu Wu$^{2}$, 
Jiarui Liu$^{2,3}$,
}
\\
\textbf{
Punya Syon Pandey$^{2,4}$, 
Roderick Wu$^{2}$, 
Chih-Hao Hsu$^{2}$,
Terry Jingchen Zhang$^{2,4}$}
\\
\textbf{
Zhijing Jin$^{1,2,4}$
,
Bernhard Schölkopf$^{1,5}$
}
\\[0.3em]
$^{1}$Max Planck Institute for Intelligent Systems, Tübingen, Germany\\
$^{2}$Jinesis AI Lab, University of Toronto \& Vector Institute \quad $^{3}$CMU \\
$^{4}$EuroSafeAI  ~~~~$^{5}$ELLIS Institute T\"ubingen
\\[0.3em]
\texttt{fdraye@tue.mpg.de}
}

\lstdefinestyle{githubcompact}{
    style=github,
    basicstyle=\ttfamily\footnotesize\color{ghidentifier},
    aboveskip=3pt,
    belowskip=3pt,
    xleftmargin=3pt,
    xrightmargin=3pt,
    lineskip=-1pt,
}

\begin{document}
\renewcommand{\thefootnote}{\fnsymbol{footnote}}
\setcounter{footnote}{1}
\maketitle
\renewcommand{\thefootnote}{\arabic{footnote}}
\setcounter{footnote}{0}
\begin{abstract}
Mechanistic interpretability seeks to understand how Large Language Models (LLMs) represent and process information. Recent approaches based on dictionary learning and transcoders enable representing model computation in terms of sparse, interpretable features and their interactions, giving rise to feature attribution graphs. However, these graphs are often large and redundant, limiting their interpretability in practice. Cross-Layer Transcoders (CLTs) address this issue by sharing features across layers while preserving layer-specific decoding, yielding more compact representations, but remain difficult to train and analyze at scale. We introduce an open-source library for end-to-end training and interpretability of CLTs. Our framework integrates scalable distributed training with model sharding and compressed activation caching, a unified automated interpretability pipeline for feature analysis and explanation, attribution graph computation using Circuit-Tracer, and a flexible visualization interface. This provides a practical and unified solution for scaling CLT-based mechanistic interpretability. Our code is available at \url{https://github.com/LLM-Interp/CLT-Forge}. 
\end{abstract}

\section{Introduction}

\begin{figure*}[t]
    \centering
    \includegraphics[width=\textwidth]{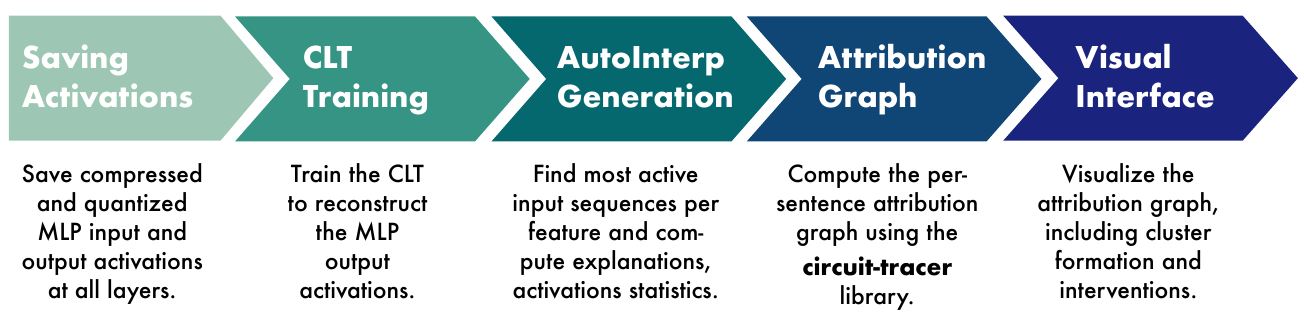}
    \caption{Overview of the CLT-Forge framework.}
    \label{fig:clt_overview}
\end{figure*}

Understanding how Large Language Models (LLMs) represent and process information is a central goal of mechanistic interpretability. A key assumption in this area is the \textit{linear representation hypothesis}, which posits that semantic features are encoded as directions in the activation space of neural networks \citep{park2023linear}. Building on this idea, \textit{transcoders} approximate the computation of MLP layers by replacing their dense nonlinear transformations with sparse, linear combinations of interpretable features computed from the same inputs \citep{dunefsky2024transcoders}. Since MLPs constitute a primary source of nonlinearity in transformer architectures, this substitution enables a structured view of model computation in terms of feature interactions. In particular, computation can be represented as a feature \textit{attribution graph}, where nodes correspond to features and edges capture causal influence through linear pathways \citep{ameisen2025circuit, lindsey2025biology}. Such graphs can be efficiently computed, pruned, analyzed, and intervened upon to yield mechanistic explanations to the internal working mechanism of LLMs~\citep{hanna2025circuit}.

However, a major limitation of this approach is that the size of attribution graphs can quickly grow to hundreds of nodes, making them difficult to analyze in practice.
This redundancy stems in part from the fact that similar features often emerge across multiple layers, leading to duplicated nodes that represent the same underlying concept in different forms.

\textit{Cross-Layer Transcoders (CLTs)} address this issue by sharing features across layers while allowing layer-specific decoding \citep{ameisen2025circuit}. This reduces redundancy and produces much more compact and interpretable attribution graphs \citep{lindsey2025landscape}. However, CLTs introduce significant computational overhead in training due to their large parameter count.

Despite recent progress, there are limited open-source frameworks that support end-to-end training of CLTs at scale, where industry labs such as Anthropic do not release their proprietary framework.
Most open-source codebases largely lack essential components such as efficient distributed training with optimal model sharding, or activation caching with quantization and compression for memory efficiency. Furthermore, post-hoc analysis remains a major bottleneck: interpreting millions of features requires retrieving activating data, computing statistics, and generating explanations (e.g., via automated interpretability), yet existing tools are fragmented and poorly integrated. Visualization is similarly limited; while tools such as Neuronpedia~\citep{linneuronpedia} provide very useful interfaces, they remain difficult to extend and integrate into research workflows.

In this work, we introduce the first unified library for CLTs that enables scalable training and post-hoc interpretability analysis. Our contributions are:

\begin{itemize}
    \item \textbf{Scalable training infrastructure}: Open-source support for large-scale CLT training with efficient GPU sharding and optimized activation caching with compression and quantization.
    \item \textbf{Efficient attribution computation}: Native integration with Circuit Tracer~\citep{hanna2025circuit} to efficiently compute and prune feature-level attribution graphs.
    \item \textbf{Built-in workflow for automated interpretability (autointerp) with flexible visual interface}: A unified and scalable autointerp pipeline with a visual interface for exploring features, visualizing attribution graphs, and performing interventions.
\end{itemize}

The code is released under the MIT License and a demonstration video is available at \url{https://youtu.be/6ptrrLawTl8}.
\section{Preliminary}

\begin{figure*}[t]
    \centering
    \includegraphics[width=\textwidth]{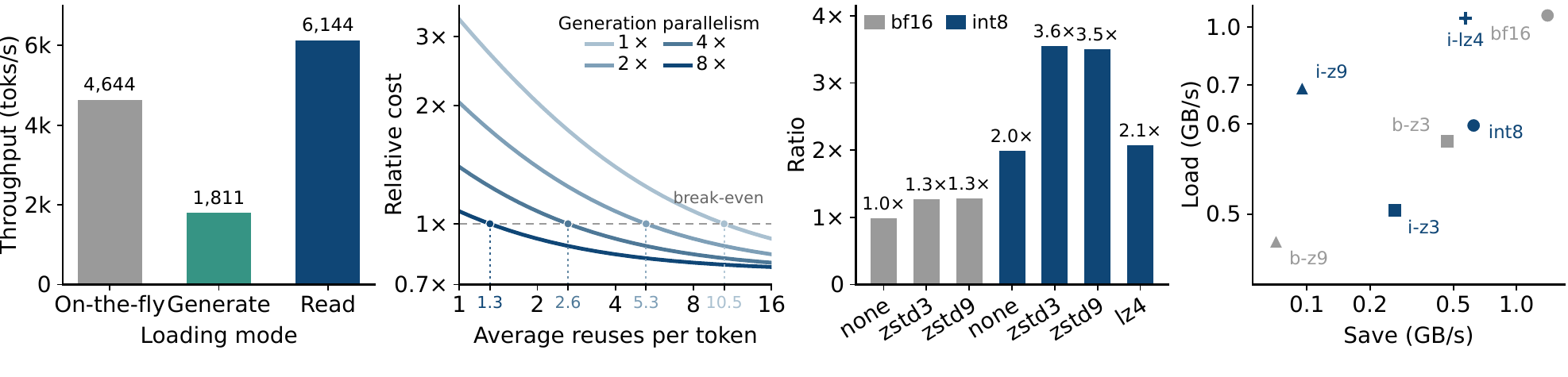}
\caption{
\textbf{Activation caching and compression trade-offs.}
\textbf{Left pair:} Activation loading throughput for on-the-fly computation, cache generation, and cached reads. Cached reads achieve the highest throughput, while cache generation incurs a one-time serialization cost. Relative training time of caching compared to on-the-fly computation as a function of the average number of reuses per activation. Curves correspond to different levels of parallel cache generation since caching can easily be parallelized, while vertical markers indicate the break-even reuse count where caching becomes faster than recomputing activations.
\textbf{Right pair:} Compression ratio achieved by different storage formats and compression methods, together with their corresponding save and load bandwidths. Int8 quantization reduces storage by up to $3.6\times$ while introducing a $12.5\%$ reconstruction error.
}
\label{fig:compression}
\end{figure*}

\subsection{Cross-Layer Transcoders}

Let $h_\ell \in \mathbb{R}^{d_{\text{model}}}$ denote the input to the MLP at layer $\ell$ for a single token position. In both standard and cross-layer transcoders, this representation is first projected into a sparse feature space via an encoder:
\begin{equation}
z_\ell = \sigma\left(W^{\ell}_{\text{enc}} h_\ell + b^{\ell}_{\text{enc}}\right) \in \mathbb{R}^{d_{\text{features}}},
\end{equation}
where $W^{\ell}_{\text{enc}} \in \mathbb{R}^{d_{\text{features}} \times d_{\text{model}}}$ and $b^{\ell}_{\text{enc}} \in \mathbb{R}^{d_{\text{features}}}$ are layer-specific parameters, and $\sigma$ denotes a sparsifying nonlinearity. In practice, we use JumpReLU as $\sigma$. In a standard transcoder, the MLP output at layer $\ell$ is reconstructed by decoding features from the same layer:
\begin{equation}
\hat{m}_\ell = W^{\ell}_{\text{dec}} z_\ell + b^{\ell}_{\text{dec}},
\end{equation}
where $W^{\ell}_{\text{dec}} \in \mathbb{R}^{d_{\text{model}} \times d_{\text{features}}}$ and $b^{\ell}_{\text{dec}} \in \mathbb{R}^{d_{\text{model}}}$.

Cross-Layer Transcoders (CLTs) extend this formulation by allowing features extracted at earlier layers to contribute to the reconstruction at later layers. The MLP output at a target layer $\ell'$ is reconstructed as:
\begin{equation}
\hat{m}_{\ell'} = \sum_{\ell \leq \ell'} W^{\ell \rightarrow \ell'}_{\text{dec}} \, z_\ell + b^{\ell'}_{\text{dec}},
\end{equation}
where $W^{\ell \rightarrow \ell'}_{\text{dec}} \in \mathbb{R}^{d_{\text{model}} \times d_{\text{features}}}$ maps features from layer $\ell$ to layer $\ell'$.

This formulation reflects the intuition that semantic features can persist across layers while undergoing transformations. For a detailed comparison between cross-layer transcoders and
standard transcoders, we refer the reader to \citep{lindsey2025landscape}. 

Cross-Layer Transcoders were introduced by Anthropic \citep{ameisen2025circuit, lindsey2025biology}, together with their training, attribution, intervention, and visualization frameworks. More recently, open-source efforts have focused on reproducing and systematically evaluating CLTs against standard transcoders \citep{lindsey2025landscape}.
Despite these advances, applications of CLTs in the literature remain limited. Existing studies include a replication of the ``greater-than'' mechanism \citep{merullo2025replicating} and a multilingual analysis \citep{harrasse2025tracing}, highlighting both the promise of CLTs and the need for more accessible and scalable infrastructure.

One existing open-source library for large-scale CLT training is Eleuther AI’s \textit{CLT-Training} fork of Sparsify \citep{eleutherai_sparsify}, which relies on on-the-fly activation computation. Additionally, their work focuses on \textit{TopK CLTs}, where only a small subset of features is active during decoding. In contrast, we do not impose activation sparsity, precluding the use of specialized kernels. Another library \citep{crosslayer-coding} also supports feature sharding along the feature dimension with flexible activation functions. However, it does not provide a unified framework integrating scalable training with automated interpretability, attribution analysis, and interactive visualization. Other libraries support CLT training for smaller-scale models (e.g., GPT-2) using Distributed Data Parallel (DDP) \citep{etredal_openclt, lange_crosslayer_transcoder}.

\subsection{Attribution Graph}
Following \citep{ameisen2025circuit}, the attribution score between feature $n$ at layer $\ell$ and position $k$, and feature $n'$ at layer $\ell'$ and position $k'$ is:
\begin{equation}
a_{\ell, k, n}^{\ell', k', n'}
=
f_{k,n}^{\ell}
\;
J_{\ell, k}^{\ell', k'}
\;
g_{k',n'}^{\ell'} .
\end{equation}

Here, $f_{k,n}^{\ell}$ denotes the decoder vector associated with feature $n$ at layer $\ell$ and position $k$, while $g_{k',n'}^{\ell'}$ denotes the corresponding encoder vector for feature $n'$ at layer $\ell'$ and position $k'$. The term $J_{\ell, k}^{\ell', k'}$ is the Jacobian mapping the MLP output at $(\ell, k)$ to the MLP input at $(\ell', k')$. This Jacobian is computed during a forward pass of the model, where nonlinearities have been frozen with a stop-gradient operation. See \citep{ameisen2025circuit} for more details. This workflow has been open-sourced in the Circuit-Tracer library \citep{hanna2025circuit}. 

\subsection{Autointerp Workflow and Visual Interface}

Once a dictionary learning model, such as an SAE or a CLT, has been trained, interpreting its features typically requires identifying highly activating sequences, computing feature statistics, and generating natural language explanations, often using LLMs \citep{bricken2023monosemanticity}.

Several libraries support parts of this workflow. Delphi \citep{paulo2024automatically} focuses on automated interpretability, \citet{sae_vis} provides feature visualization, and Neuronpedia offers a comprehensive interface for feature exploration and attribution analysis \citep{linneuronpedia}. However, these tools address different stages of the workflow and are not tightly integrated, often requiring practitioners to combine multiple libraries. This motivates a unified framework that integrates scalable automated interpretability, attribution analysis, and an extensible visualization interface.

\begin{figure*}[t]
    \centering
    \includegraphics[width=\textwidth]{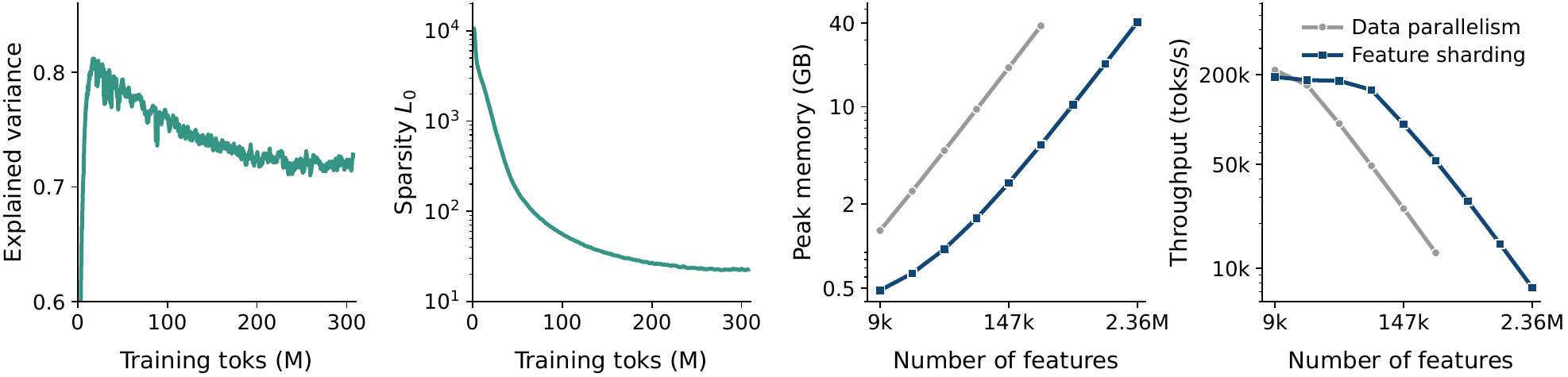}
\caption{
\textbf{CLT-Forge training dynamics and scalability.}
\textbf{Left pair:} Training dynamics of a Cross-Layer Transcoder on Llama-3.2-1B activations with 65,536 features per layer (16 layers), trained on 300M OpenWebText tokens in 17 hours using a single 8$\times$H100 node. To our knowledge, this is the first open-source implementation for training CLTs on a 1B-scale language model. 
\textbf{Right pair:} Comparison of feature sharding and data parallelism as the total number of features increases. Feature sharding dramatically reduces per-GPU memory while improving throughput, enabling substantially larger dictionaries than conventional data-parallel training.
}\label{fig:training_dynamics}
\end{figure*}

\section{CLT-Forge: Design and features}

We describe in the following subsections the different components of the pipeline as described in Figure~\ref{fig:clt_overview}. 

\subsection{Activation Caching}

\label{sec:caching}

Training a CLT requires reconstructing MLP outputs from their corresponding inputs, which in turn necessitates access to activations at all layers over large-scale datasets (e.g., hundreds of millions of tokens). We support two modes for handling activations. First, activations can be computed \textit{on the fly}, which avoids storing large activation datasets and is useful in disk memory-constrained settings or during rapid experimentation (e.g., when exploring architectures or hyperparameters) \citep{bloom2024saetrainingcodebase}. The second approach is to precompute and store activations on disk prior to CLT training, which is more efficient in practice \citep{ameisen2025circuit}. We compare the trade-offs between on-the-fly activation computation and cached activations in Figure~\ref{fig:compression}. Because activation caching can be generated independently across multiple workers, the one-time preprocessing cost is quickly amortized, making cached activations substantially more efficient once activations are reused multiple times during training.

However, activation caching typically requires several terabytes of storage, making naive caching impractical. CLT-Forge therefore supports a range of quantization and compression schemes, including symmetric per-layer int8 quantization together with optional zstd and LZ4 compression. Activations are stored in independently compressed fixed-size chunks together with their per-layer scale factors, enabling efficient streaming during training. Rather than providing a single compression mode, the framework exposes a trade-off between storage footprint and I/O performance, as shown in Figure~\ref{fig:compression}. Lightweight compression (e.g., LZ4) prioritizes fast writes and reads, while stronger entropy coding (e.g., zstd) achieves higher compression ratios at the expense of slower cache generation. Since activation caches are typically written once but reused over many training epochs, this one-time write overhead is often amortized, making higher compression attractive in practice. In our experiments, int8 quantization with zstd reduces storage by up to $3.6\times$ while maintaining high read throughput, reducing the activation cache of a 300M-token Llama-3.2-1B CLT training run from approximately 20\,TB to around 4\,TB.

During training, activation chunks are loaded into memory as needed. In a Distributed Data Parallel (DDP) setting, each worker processes different chunks. In contrast, when the CLT is sharded along the feature dimension (see Section~\ref{sec:sharding}), all workers operate on the same data but on different subsets of features. Finally, similar to \citep{bloom2024saetrainingcodebase}, we normalize activations to have approximately unit norm by dividing each activation by a normalization factor estimated from the first $N$ batches.

\subsection{Sparse Decode}

We use JumpReLU as the activation and support multiple sparsity schedulers,
following the initialization and training recommendations from
\citep{anthropic2025janupdate}. We also support TopK.

For cross-layer CLTs, the decoder $z W_\text{dec}^\top$ dominates the
per-step FLOPs: each active latent writes to every target layer, so the cost
scales with the full expansion $K$ regardless of how sparse $z$ actually is.
This is wasted work as soon as training pushes activations into the sparse
regime, for the Llama 1B CLT in Figure~\ref{fig:training_dynamics}, $L_0$ has
already dropped below $100$ by a third of the way through training, and keeps
falling.

We exploit this by adding a sparse-aware decode path built on
\texttt{torch.nn.functional.embedding\_bag}: for each token we look up only the
active rows of $W_\text{dec}$ and sum them in a single fused kernel, replacing
the dense $\mathcal{O}(K\,d_\text{in})$ matmul with an
$\mathcal{O}(k\,d_\text{in})$ gather where $k=\lVert z\rVert_0 \ll K$. 

\begin{figure}
    \centering
    \includegraphics[width=0.90\linewidth]{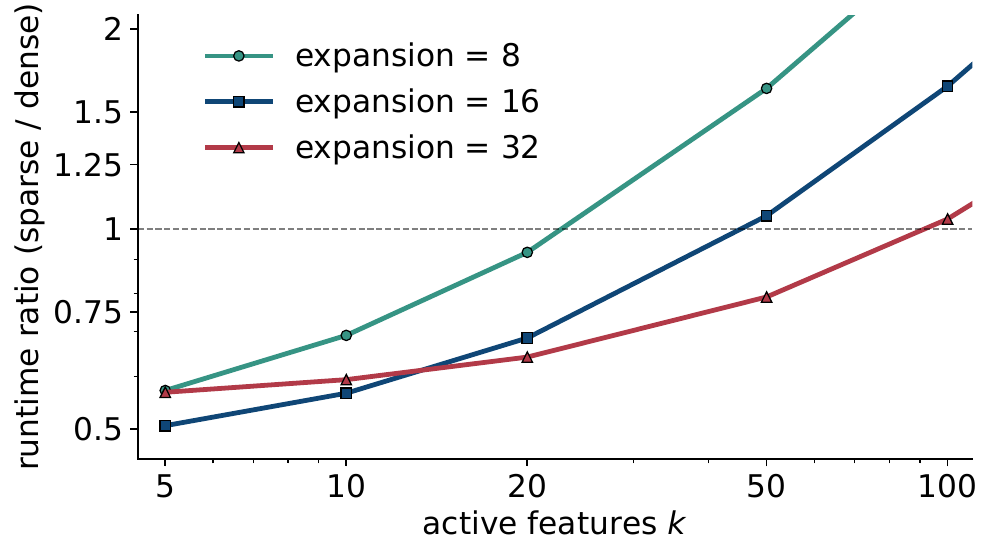}
    \caption{Sparse (\texttt{embedding\_bag}) vs.\ dense decode runtime ratio at
    fixed $k$ (bf16, $d_\text{in}=2048$, cross-layer JumpReLU CLT). Below 1.0,
    sparse wins; the break-even $k$ scales roughly linearly with expansion —
    ${\sim}20$, $50$, $100$ for $8\times$, $16\times$, $32\times$.}
    \label{fig:sparse-decode}
\end{figure}

Figure~\ref{fig:sparse-decode} shows that the crossover $k$ grows with
expansion, so wider dictionaries benefit even more from sparse decoding. At runtime we monitor
per-batch $L_0$ and auto-switch from the dense to the sparse path once
sparsity crosses the crossover threshold, which, in practice, happens well
inside training and remains true at inference.

\subsection{GPU Sharding}
\label{sec:sharding}

As model size scales, the corresponding CLTs also quickly become too large for a single GPU.
While Distributed Data Parallel (DDP) could suffice for training CLTs on smaller models such as GPT-2, scaling to larger models would inevitably face memory limitations.

Although Fully Sharded Data Parallel (FSDP) could help in this front, it remains suboptimal for CLT architectures and encounters memory constraints at larger scales.
Following~\citep{ameisen2025circuit}, we instead adopt feature sharding, where features are partitioned across GPUs and the partial outputs are aggregated at the end of the forward pass. As shown in Figure~\ref{fig:training_dynamics}, feature sharding dramatically reduces per-GPU memory while improving throughput, enabling substantially larger Cross-Layer Transcoders than conventional data parallelism. Using this implementation, we train a Cross-Layer Transcoder on Llama-3.2-1B with 65,536 features per layer (16 layers) over 300M OpenWebText tokens in 17 hours on a single 8$\times$H100 node. To our knowledge, this is the first open-source implementation supporting CLT training on a 1B-scale language model. 
\begin{figure*}[t]
    \centering
    \includegraphics[width=\textwidth]{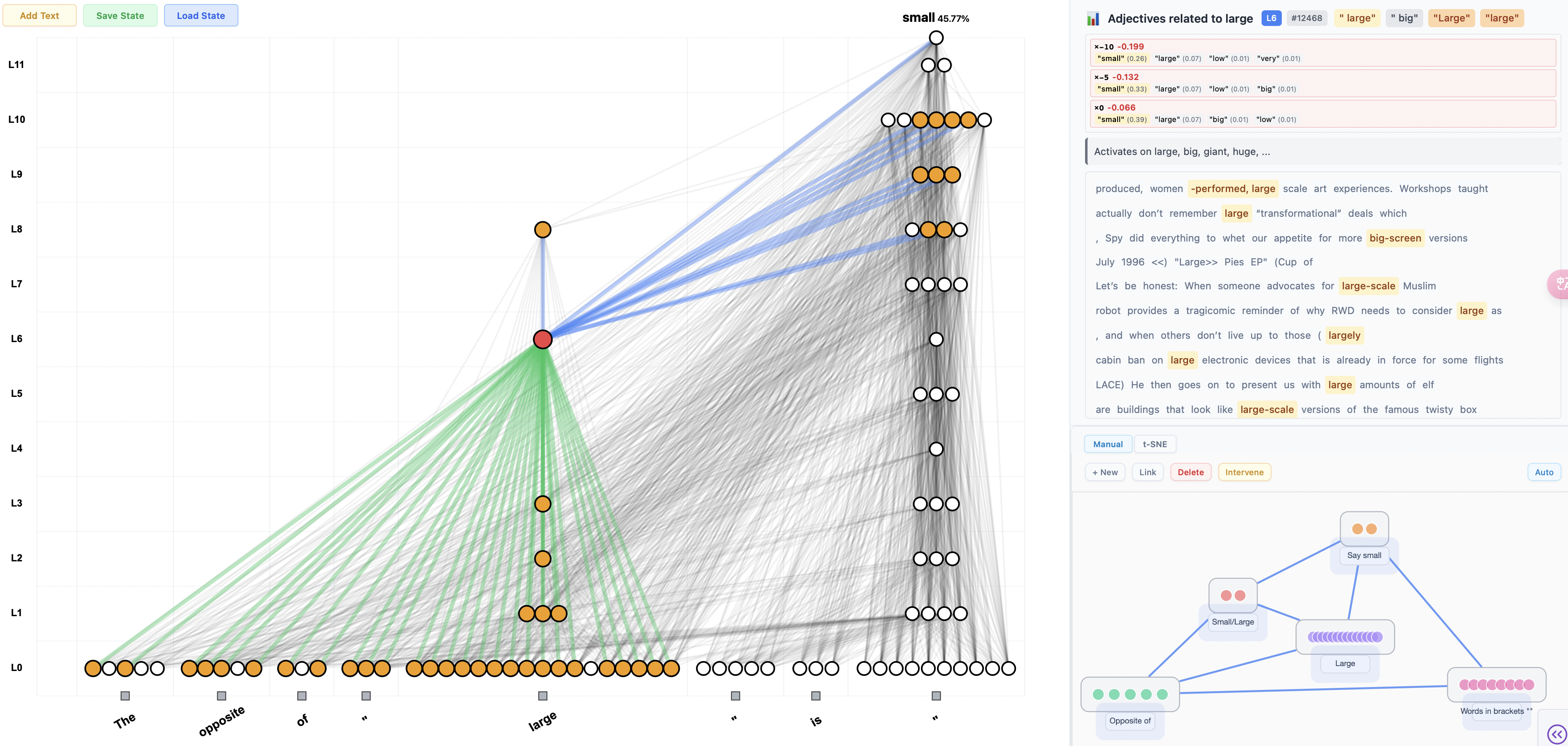}
    \caption{Example of a GPT-2 attribution-graph for the sentence 'The opposite of "large" is' with replacement-score $0.77$.}
    \label{fig:visual_interface}
\end{figure*}

\subsection{Automated Interpretability }

For automated interpretability, we parallelize computation over the feature dimension. Each independent worker maintains a subset of encoder vectors on the GPU and computes the corresponding latent activations over large token streams (e.g., 10M tokens). During activation loading, we incrementally track and update the top-$K$ activating sequences directly on the GPU, ensuring that memory usage remains minimal beyond storing these top activations. Within the same pass, we construct feature-level summaries, including top activating tokens, sequences, and optional metadata (e.g., language). If required by the user, these summaries are then used to automatically generate prompts for LLM-based explanation, providing representative examples and instructions to produce coherent feature descriptions. Results are finally stored in a per-worker database keyed by layer and feature index, which are merged into a single unified store upon completion, reducing the cost of writing millions of individual files. This design enables a fully integrated and parallelized, single-pass autointerp pipeline. 

\subsection{Attribution Graph}

For the attribution graph computation, we map our CLT to Circuit-Tracer~\citep{hanna2025circuit}, which allows us to compute, prune the attribution graph and perform interventions when needed. 

\subsection{Visual Interface}

We developed a Python-based Dash interface (Figure~\ref{fig:visual_interface}) that enables interactive visualization and exploration of attribution graphs.
The interface provides feature-level information, including top activating tokens, representative sequences, automatically generated explanations, and interventions. 

In addition, similar to \citep{linneuronpedia}, the interface includes tools for clustering features and constructing higher-level structures. Users can define clusters, connect them via edges, and perform interventions at the cluster level. These operations trigger background jobs that leverage Circuit Tracer to compute the corresponding effects efficiently~\citep{hanna2025circuit}. The main objective of this interface is to provide a simplified and more easily extensible alternative to \citep{linneuronpedia}, enabling rapid experimentation and iteration in academic settings.

\section{Evaluation}

\paragraph{Evaluation.}
As CLT-Forge is a systems library rather than a new modeling method, we evaluate it along two complementary axes: correctness and scalability. Across both GPT-2 and Llama-3.2-1B, CLT-Forge reproduces the behavior of existing closed-source Cross-Layer Transcoder implementations, achieving approximately 0.7-0.8 explained variance, 0.8 replacement score, and 0.95 graph completeness at low sparsity levels \citep{hanna2025circuit, merullo2025replicating, lindsey2025landscape, harrasse2025tracing}. We then evaluate the systems components that enable large-scale training, including feature sharding, activation caching, and compression, and demonstrate training of a 65,536-feature-per-layer Cross-Layer Transcoder on Llama-3.2-1B over 300M OpenWebText tokens in 17 hours on a single $8\times$H100 node. Finally, we illustrate the complete workflow through end-to-end examples of automated interpretation, attribution analysis, and interactive visualization.

\section{Limitations and Future Work}
In this section, we outline several key directions for future work, many of which reflect open problems in mechanistic interpretability at large.
First, attention maps are currently frozen when computing attribution graphs; incorporating attention attribution is an important next step~\citep{kamath2025tracing}. We plan to use the approach of~\citep{draye2025sparse}.
Second, CLTs remain expensive to train and analyze due to their parameter count and memory footprint, motivating research on more compute- and memory-efficient architectures.
Third, reconstruction error is optimized as a proxy for replacement score~\citep{lindsey2025landscape}. We also found that directly optimizing replacement score led to unstable and noisy results, suggesting the need for more robust objectives or optimization schemes.

\section{Conclusion}

We introduced CLT-Forge, the first unified open-source framework for scalable training and interpretability of Cross-Layer Transcoders. Our library integrates efficient feature-sharded training, compressed activation caching, automated interpretability, attribution graph computation, and an extensible visualization interface into a single workflow. By combining these components, CLT-Forge enables end-to-end analysis of CLTs at scales that were previously difficult to access in open-source settings. We hope that this framework lowers the barrier to studying mechanistic interpretability at scale, facilitates reproducible research, and supports rapid experimentation in academic environments. Future work will focus on improving efficiency, extending attribution methods (e.g., attention attribution), and open-sourcing CLTs for larger models.

\section*{Acknowledgement}

This material is based in part upon work supported by the German Federal Ministry of Education and Research (BMBF): Tübingen AI Center, FKZ: 01IS18039B; by the Machine Learning Cluster of Excellence, EXC number 2064/1 – Project number 390727645; by Schmidt Sciences SAFE-AI Grant; by Coefficient Giving; and
by the Survival and Flourishing Fund.
Resources used in preparing this research project were provided, in part, by the Province of Ontario, the Government of Canada through CIFAR, and companies sponsoring the Vector Institute.
F. D. acknowledges support through a fellowship from the Hector Fellow Academy.

\bibliography{custom}

\appendix
\onecolumn

\section{Usage Examples}
\label{sec:usage_examples}

Minimal examples for the main CLT-Forge workflow are shown below. Complete
configurations and executable scripts are available in the repository.

\vspace{5pt}

\noindent\textbf{Activation caching.}
\vspace{2pt}

\begin{lstlisting}[style=githubcompact]
from clt_forge import load_model, ActivationsStore
from clt_forge import CLTTrainingRunnerConfig

model = load_model("Llama-3.2-1B")
cfg = CLTTrainingRunnerConfig(expansion_factor=48)
store = ActivationsStore(model, cfg)
store.generate_and_save_activations()
\end{lstlisting}
\vspace{5pt}

\noindent\textbf{Training.}
\vspace{2pt}
\begin{lstlisting}[style=githubcompact]
from clt_forge import CLTTrainingRunner

CLTTrainingRunner(cfg).run()
\end{lstlisting}
\vspace{5pt}

\noindent\textbf{Automated interpretability.}
\vspace{2pt}
\begin{lstlisting}[style=githubcompact]
from clt_forge import AutoInterp, AutoInterpConfig

cfg = AutoInterpConfig(
    model_name="Llama-3.2-1B",
    clt_path="path/to/checkpoint",
)
AutoInterp(cfg).run("path/to/features")
\end{lstlisting}
\vspace{5pt}

\noindent\textbf{Attribution graphs.}
\vspace{2pt}
\begin{lstlisting}[style=githubcompact]
from clt_forge import AttributionRunner

runner = AttributionRunner(
    model_name="Llama-3.2-1B",
    clt_path="path/to/checkpoint",
)
graph = runner.run("The opposite of 'large' is")
\end{lstlisting}
\vspace{5pt}

\noindent\textbf{Visualization.}
\vspace{2pt}
\begin{lstlisting}[style=githubcompact]
from clt_forge import launch_interface

launch_interface(graph, "path/to/features")
\end{lstlisting}

\section{Additional Training Details}
\label{sec:additional_training_details}

\paragraph{Training objective.}
Following prior work, we train the CLT to reconstruct the MLP outputs while
encouraging sparse feature activations and penalizing dead features. The
overall objective is
\begin{align}
\mathcal{L}
={}&
\underbrace{
\sum_{\ell'}
\left\|
\hat{m}_{\ell'} - m_{\ell'}
\right\|_2^2
}_{\text{reconstruction}}
\nonumber +
\lambda_0
\underbrace{
\sum_{\ell}
\tanh\!\left(
C\,
z_\ell \odot
\left\|
\mathbf{W}_{\mathrm{dec}}^\ell
\right\|
\right)
}_{\text{sparsity}}
\nonumber +
\lambda_1
\underbrace{
\sum_{\ell}
\mathrm{ReLU}\!\left(
\exp(\tau)-h_\ell^{\mathrm{pre}}
\right)
\left\|
\mathbf{W}_{\mathrm{dec}}^\ell
\right\|
}_{\text{dead-feature penalty}} 
\end{align}
Here, $\mathbf{W}_{\mathrm{dec}}^\ell$ denotes the concatenation of all
decoder weights originating from layer $\ell$,
$h_\ell^{\mathrm{pre}}$ denotes the feature pre-activations,
$\tau$ is the dead-feature threshold, and $C$ is a scaling constant.
The coefficients $\lambda_0$ and $\lambda_1$ control the sparsity and
dead-feature regularization terms, respectively.

\paragraph{Parameter scaling.}
A CLT associates each transformer layer with an encoder
$\mathbb{R}^{d} \rightarrow \mathbb{R}^{ed}$ and each ordered layer pair
$(\ell_s,\ell_t)$ with $\ell_s < \ell_t$ with a decoder
$\mathbb{R}^{ed} \rightarrow \mathbb{R}^{d}$, where
$ed=d_{\mathrm{features}}$. Ignoring biases, the total number of parameters is
\begin{equation}
N_{\mathrm{CLT}}(L,d,e)
=
\left(
\frac{L(L-1)}{2}+L
\right) e d^2 .
\end{equation}
The parameter count therefore grows quadratically with both the number of
layers and the model dimension. For Llama-3.2-1B, with $L=16$, $d=2048$,
and $e=48$, this gives approximately $27.4$B parameters, motivating the
feature-sharded training strategy used in CLT-Forge.

\section{Configuration}

\begin{table}[h]
    \centering
    \small
    \setlength{\tabcolsep}{5pt}
    \renewcommand{\arraystretch}{1.12}
    \begin{threeparttable}
    \caption{
        Optimal training configurations for the GPT-2 and Llama-3.2-1B
        Cross-Layer Transcoders (CLTs).
    }
    \label{tab:clt_training_config}
    \begin{tabularx}{\linewidth}{
        >{\raggedright\arraybackslash}X
        >{\centering\arraybackslash}p{0.27\linewidth}
        >{\centering\arraybackslash}p{0.27\linewidth}
    }
        \toprule
        \textbf{Configuration} &
        \textbf{GPT-2} &
        \textbf{Llama-3.2-1B} \\
        \midrule

        Dictionary size, $d_{\mathrm{latent}}$
            & 24,576
            & 65,536 \\
        Expansion factor
            & $32\times$
            & $32\times$ \\

        \addlinespace
        Training tokens
            & 209M
            & 307M \\
        Token batch size
            & 8192
            & 8192 \\
        Adam $(\beta_1,\beta_2)$
            & $(0.9,\,0.999)$
            & $(0.9,\,0.999)$ \\
        Peak learning rate
            & $2\times 10^{-4}$
            & $2\times 10^{-4}$ \\
        Sparsity coefficient, $\lambda_0$
            & 2
            & 12 \\
        Numerical precision
            & FP32\tnote{a}
            & BF16 \\

        \addlinespace
        Hardware
            & $8\times$ H100 80GB
            & $8\times$ H100 80GB \\
        Wall-clock training time
            & 5h 04m
            & 17h 35m \\
        Final mean $L_0$
            & 10.0
            & 22.51 \\
        Mean explained variance
            & 0.75
            & 0.746 \\

        \bottomrule
    \end{tabularx}
    \end{threeparttable}
\end{table}

\section{Example of Training Dynamics}
\label{sec:appendix_dynamics}

\begin{figure}[h]
    \centering
    \includegraphics[width=1.0\linewidth]{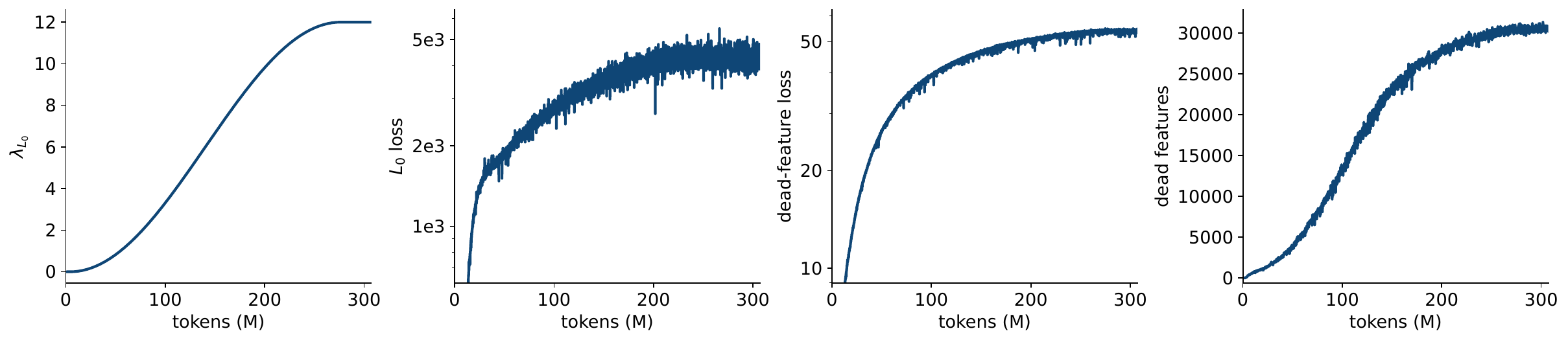}
    \caption{Training curves for the Llama-3.2-1B CLT over ~307 M tokens. Left to right: L0-penalty coefficient $\lambda_{L_0}$ (warmup schedule), $L_0$ loss, dead-feature auxiliary loss, and dead-feature count. Log-y axes are clipped to the post-warmup range (from 5\% of tokens) so the initial spike is off-frame.}
    \label{fig:placeholder}
\end{figure}

\begin{figure}[h]
    \centering
    \includegraphics[width=1.0\linewidth]{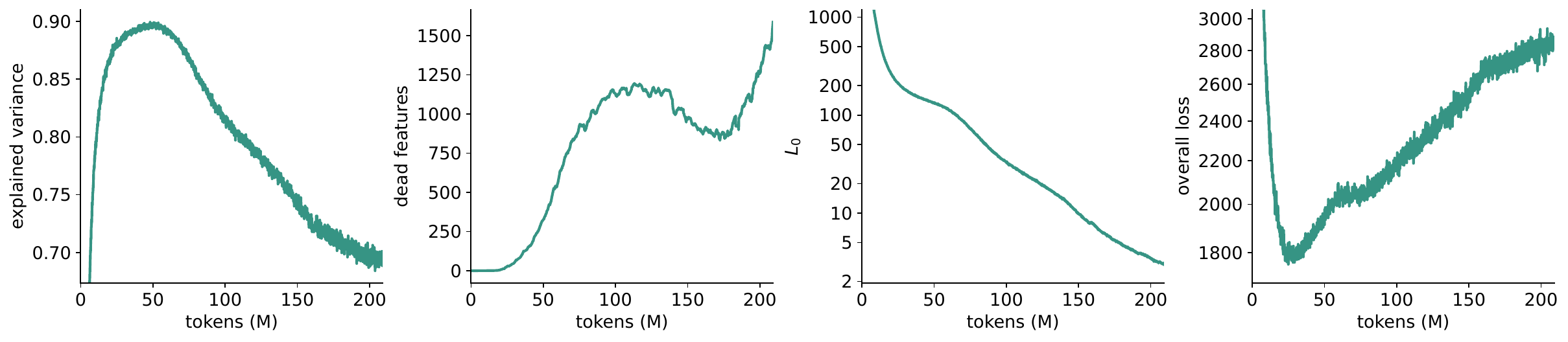}
    \caption{Training curves for the GPT-2 CLT over ~209 M tokens. Left to right: explained variance on the reconstruction, dead-feature count, average $L_0$ per token, and overall training loss. As $\lambda_{L_0}$ ramps up, $L_0$ collapses by three decades and dead features grow, while explained variance peaks early and then degrades as the sparsity penalty tightens.}
    \label{fig:placeholder}
\end{figure}

\end{document}